\documentclass[11pt]{article}

\usepackage[utf8]{inputenc}
\usepackage[margin=1in]{geometry}
\usepackage{amsmath,amssymb,amsfonts}
\usepackage{graphicx}
\usepackage{url}
\usepackage{hyperref}
\usepackage{cite}

\title{Trustworthy LLM-Mediated Communication: Evaluating Information Fidelity in LLM as a Communicator (LAAC) Framework in Multiple Application Domains}

\author{
Mohammed Musthafa Rafi, Adarsh Krishnamurthy, Aditya Balu\\
\textit{Iowa State University}\\
Ames, IA, USA\\
\texttt{\{mohd7, adarsh, baditya\}@iastate.edu}
}

\date{}

\begin{document}

\maketitle

\begin{abstract}
The proliferation of AI-generated content has created an absurd communication theater where senders use LLMs to inflate simple ideas into verbose content, recipients use LLMs to compress them back into summaries, and as a consequence neither party engage with authentic content. LAAC (LLM as a Communicator) proposes a paradigm shift—positioning LLMs as intelligent communication intermediaries that capture the sender's intent through structured dialogue and facilitate genuine knowledge exchange with recipients. Rather than perpetuating cycles of AI-generated inflation and compression, LAAC enables authentic communication across diverse contexts including academic papers, proposals, professional emails, and cross-platform content generation. However, deploying LLMs as trusted communication intermediaries raises critical questions about information fidelity, consistency, and reliability. This position paper systematically evaluates the trustworthiness requirements for LAAC's deployment across multiple communication domains. We investigate three fundamental dimensions: (1) Information Capture Fidelity—accuracy of intent extraction during sender interviews across different communication types, (2) Reproducibility—consistency of structured knowledge across multiple interaction instances, and (3) Query Response Integrity—reliability of recipient-facing responses without hallucination, source conflation, or fabrication. Through controlled experiments spanning multiple LAAC use cases, we assess these trust dimensions using LAAC's multi-agent architecture. Preliminary findings reveal measurable trust gaps that must be addressed before LAAC can be reliably deployed in high-stakes communication scenarios.
\end{abstract}


\section{Introduction}

The rapid adoption of Large Language Models (LLMs) has fundamentally transformed how people create and consume written content. However, this transformation has produced an unexpected and inefficient phenomenon: a communication theater where authenticity is systematically eliminated at both ends of the exchange. Senders prompt LLMs to expand terse bullet points into elaborate prose, adding formality and verbosity that obscures rather than clarifies their original intent. Recipients, overwhelmed by this inflated content, immediately prompt their own LLMs to compress these expansive texts back into bullet-point summaries. The result is a complete cycle of artificial inflation and deflation where genuine human thought is lost in translation, computational resources are wasted, and neither party engages meaningfully with the actual content.

This wasteful cycle affects nearly every domain of written communication—from academic papers artificially expanded with filler text to business emails that say in five paragraphs what could be expressed in two sentences. The fundamental problem is not the LLM technology itself, but rather its misapplication as a content generator rather than as an intelligent intermediary. We argue that LLMs should not replace human communication but rather facilitate it—serving as trusted mediators that faithfully capture sender intent and enable recipient comprehension.

We propose LAAC (LLM as a Communicator), a paradigm that repositions LLMs from content generators to communication intermediaries. Rather than producing verbose expansions of simple ideas, LAAC systems engage senders in structured dialogue to extract and formalize their authentic intent, then enable recipients to interact directly with this structured knowledge through natural language queries. This approach promises to restore authenticity to LLM-mediated communication while leveraging the genuine strengths of these systems—their ability to facilitate knowledge structuring, semantic understanding, and interactive question-answering.

However, deploying LLMs as communication intermediaries introduces critical trustworthiness challenges. If an LLM misrepresents a sender's intent, produces inconsistent knowledge structures from identical inputs, or fabricates information when responding to recipient queries, the entire communication exchange becomes unreliable. Unlike traditional LLM applications where errors might be caught and corrected by human review, LAAC systems operate as autonomous intermediaries where trust is essential for adoption.

This position paper makes the following contributions:
\begin{itemize}
\item We identify and formalize three fundamental trust dimensions for LAAC systems: Information Capture Fidelity, Reproducibility, and Query Response Integrity
\item We present a systematic evaluation methodology for assessing these trust dimensions across diverse communication domains
\item We describe a concrete implementation of LAAC for academic paper development and review, demonstrating the framework's applicability to high-stakes communication scenarios
\item We report preliminary findings from controlled experiments that reveal measurable gaps in current LLM trustworthiness for communication intermediary roles
\item We propose future research directions for improving LAAC system reliability and establishing verifiable trust metrics
\end{itemize}

The remainder of this paper is structured as follows. Section \ref{sec:related} reviews related work in LLM-mediated communication and trustworthy AI systems. Section \ref{sec:architecture} details the LAAC framework architecture and its three-agent design. Section \ref{sec:methodology} presents our evaluation methodology for measuring trustworthiness across the three critical dimensions. Section \ref{sec:implementation} describes our implementation in the academic paper domain. Section \ref{sec:findings} discusses preliminary findings and their implications. Section \ref{sec:conclusion} concludes with future research directions.

\section{Related Work}
\label{sec:related}

\subsection{LLM-Mediated Communication}

The use of LLMs to assist in various forms of written communication has become widespread, with applications ranging from email composition \cite{email_assist} to academic writing support \cite{academic_writing}. However, most existing systems focus on content generation rather than communication mediation. Tools like GPT-4 \cite{gpt4}, Claude \cite{claude}, and specialized writing assistants \cite{grammarly} typically operate in a one-directional manner—transforming user inputs into polished outputs without maintaining semantic fidelity or enabling recipient interaction with the underlying intent.

Recent work has explored more interactive approaches to AI-assisted writing. Collaborative writing systems \cite{collaborative_ai} enable iterative refinement through dialogue, but still position the LLM as a co-author rather than a faithful intermediary. Similarly, document querying systems \cite{doc_qa} allow users to ask questions about existing documents, but these systems analyze pre-written content rather than capturing intent during document creation.

\subsection{Information Fidelity and Hallucination}

A critical challenge for deploying LLMs in high-stakes applications is their tendency to generate plausible but factually incorrect information—a phenomenon known as hallucination \cite{hallucination}. Extensive research has documented LLM hallucinations across domains including factual question-answering \cite{factuality}, mathematical reasoning \cite{math_reasoning}, and citation generation \cite{citation_hallucination}.

Efforts to mitigate hallucination include retrieval-augmented generation \cite{rag}, fact-checking mechanisms \cite{factcheck}, and uncertainty quantification \cite{uncertainty}. However, these approaches primarily address factual accuracy in external knowledge domains rather than fidelity to user-provided information—the core requirement for communication intermediary systems.

\subsection{Multi-Agent LLM Systems}

Recent advances in multi-agent LLM architectures have demonstrated improved performance through task decomposition and specialized agent roles \cite{multiagent}. Systems like AutoGPT \cite{autogpt} and MetaGPT \cite{metagpt} employ multiple agents with distinct responsibilities to accomplish complex tasks. These architectures inspire LAAC's design, but existing multi-agent systems focus on task completion rather than communication mediation.

\subsection{Trustworthy AI and Verification}

The broader field of trustworthy AI has established principles for reliable system deployment including transparency, explainability, fairness, and robustness \cite{trustworthy_ai}. Verification methods for neural systems \cite{verification} provide formal guarantees for specific properties, but these techniques have not been extensively applied to communication fidelity scenarios.

Our work bridges these research areas by applying trustworthiness principles specifically to LLM-mediated communication, proposing concrete metrics for information fidelity, and presenting an evaluation methodology suitable for communication intermediary systems.

\section{LAAC Framework Architecture}
\label{sec:architecture}

The LAAC (LLM as a Communicator) framework employs a multi-agent architecture designed to maintain information fidelity throughout the communication pipeline. Unlike traditional LLM applications where a single model attempts to serve all functions, LAAC decomposes the communication mediation task into three specialized agent roles, each optimized for a specific trust-critical function.

\subsection{Three-Agent Architecture}

\subsubsection{Interview Agent}

The Interview Agent serves as the primary interface between the sender and the LAAC system. Its role is to extract comprehensive information about the sender's intent through structured dialogue. Rather than accepting free-form text and attempting to expand it, the Interview Agent asks targeted questions that elicit specific details about the sender's message.

For academic paper development, the Interview Agent asks about research methodology, key findings, theoretical contributions, and related work. For business communications, it might probe for objectives, action items, stakeholders, and deadlines. The Interview Agent's prompts are domain-specific and designed to capture all information necessary for complete knowledge representation.

Critical to trustworthiness, the Interview Agent's objective is information extraction, not content generation. It should ask questions rather than make assumptions, request clarification when sender responses are ambiguous, and explicitly acknowledge gaps in understanding rather than filling them with inferred content.

\subsubsection{Extraction Agent}

The Extraction Agent processes the complete interview transcript and generates a structured knowledge representation of the sender's intent. This representation serves as the canonical source of truth for all subsequent interactions—it is what the sender actually meant to communicate.

The knowledge structure varies by domain but generally includes hierarchical organization, explicit relationships between concepts, supporting evidence or reasoning for claims, and metadata about certainty levels. For academic papers, this might be organized by standard paper sections (introduction, methodology, results, discussion). For other communications, alternative structures may be appropriate.

The Extraction Agent's trustworthiness depends on its ability to accurately reflect the interview content without introducing new information, misrepresenting relationships, or omitting critical details. Its output must be a faithful transformation of the sender's expressed intent into a queryable format.

\subsubsection{Query Agent}

The Query Agent provides the recipient interface to the structured knowledge. Recipients can ask natural language questions about any aspect of the sender's message, and the Query Agent responds based solely on the extracted knowledge structure. This enables recipients to efficiently access relevant information without reading verbose content while maintaining fidelity to the sender's intent.

The Query Agent faces perhaps the most critical trustworthiness challenge: it must resist the strong LLM tendency to hallucinate. When faced with questions that cannot be answered from the structured knowledge, it must explicitly acknowledge this limitation rather than generating plausible but unfounded responses. It must also avoid conflating information from different sections or misrepresenting the certainty of claims.

\subsection{Information Flow}

The complete LAAC communication flow proceeds as follows:

\begin{enumerate}
\item The sender initiates communication by engaging with the Interview Agent
\item The Interview Agent conducts a structured dialogue, asking domain-specific questions to extract comprehensive information about the sender's intent
\item The complete interview transcript is processed by the Extraction Agent, which generates a structured knowledge representation
\item The sender reviews and potentially refines the extracted knowledge structure
\item Recipients interact with the Query Agent, asking questions that are answered based on the structured knowledge
\item The Query Agent responds to recipient questions by retrieving and presenting relevant information from the knowledge structure
\end{enumerate}

This architecture ensures that the sender's intent is captured once, represented faithfully, and accessed consistently by all recipients. No content inflation or compression occurs—only information extraction, structuring, and retrieval.

\subsection{Domain Adaptability}

While this paper focuses primarily on academic paper development as a demonstration domain, the LAAC architecture is designed for adaptability across communication contexts. The three-agent structure remains constant, but agent prompts, knowledge schemas, and interaction patterns are customized for each domain:

\begin{itemize}
\item \textbf{Academic Papers}: Interview about methodology, results, contributions; structure as paper sections; enable detailed technical queries
\item \textbf{Business Proposals}: Interview about objectives, solutions, timelines; structure as proposal components; enable stakeholder-specific queries
\item \textbf{Professional Emails}: Interview about context, action items, dependencies; structure as topic hierarchy; enable quick information lookup
\item \textbf{Technical Documentation}: Interview about system architecture, APIs, use cases; structure as technical reference; enable developer queries
\end{itemize}

This domain adaptability is essential for LAAC's practical utility, but it also multiplies the trustworthiness evaluation challenge—each domain may exhibit different failure modes and require domain-specific trust metrics.

\section{Trustworthiness Evaluation Methodology}
\label{sec:methodology}

We propose a systematic methodology for evaluating LAAC trustworthiness across three fundamental dimensions. Each dimension addresses a critical aspect of the communication intermediary role and requires distinct evaluation approaches.

\subsection{Dimension 1: Information Capture Fidelity}

Information Capture Fidelity measures how accurately the Interview Agent extracts information from the sender and how faithfully the Extraction Agent represents this information in the knowledge structure. This dimension addresses the fundamental question: Does the structured knowledge accurately reflect what the sender intended to communicate?

\subsubsection{Evaluation Protocol}

We employ a ground-truth comparison methodology:
\begin{enumerate}
\item Select representative content samples from the target domain (e.g., existing academic papers, business proposals)
\item For each sample, conduct an Interview Agent session where a domain expert role-plays as the author, providing information based on the original document
\item Generate the extracted knowledge structure
\item Compare the knowledge structure against the original document using both automated metrics and human evaluation
\end{enumerate}

\subsubsection{Metrics}

We assess fidelity across multiple dimensions:
\begin{itemize}
\item \textbf{Content Coverage}: Percentage of key concepts from the original appearing in the structured knowledge
\item \textbf{Semantic Accuracy}: Correctness of relationships and claims in the knowledge structure (human-evaluated)
\item \textbf{Information Addition}: Percentage of concepts in the structured knowledge that do not appear in the interview or original (false positives)
\item \textbf{Information Omission}: Percentage of key concepts from the interview that do not appear in the structured knowledge (false negatives)
\end{itemize}

\subsection{Dimension 2: Reproducibility}

Reproducibility measures the consistency of the LAAC system when processing the same sender intent multiple times. Given identical or near-identical interview inputs, does the system produce equivalent knowledge structures? This dimension is critical for establishing LAAC as a reliable communication medium.

\subsubsection{Evaluation Protocol}

We employ a repeated-extraction methodology:
\begin{enumerate}
\item Conduct an Interview Agent session on a specific topic
\item Process the same interview transcript through the Extraction Agent multiple times (varying random seeds, temperature parameters)
\item Compare the resulting knowledge structures for consistency
\item Alternatively, conduct multiple interview sessions with the same information and compare extracted structures
\end{enumerate}

\subsubsection{Metrics}

We measure consistency using:
\begin{itemize}
\item \textbf{Structural Similarity}: Overlap in hierarchical organization and section structure
\item \textbf{Semantic Equivalence}: Agreement on core claims and relationships (evaluated via semantic similarity metrics)
\item \textbf{Detail Consistency}: Stability of specific facts, numbers, and citations across extractions
\item \textbf{Variability Analysis}: Quantification of differences and identification of factors causing inconsistency
\end{itemize}

\subsection{Dimension 3: Query Response Integrity}

Query Response Integrity assesses the trustworthiness of the Query Agent's responses to recipient questions. This dimension addresses whether recipients can rely on the information provided by the system, focusing specifically on hallucination avoidance, source fidelity, and appropriate uncertainty acknowledgment.

\subsubsection{Evaluation Protocol}

We employ a question-answer verification methodology:
\begin{enumerate}
\item Generate a knowledge structure from a known ground-truth document
\item Create a test set of questions spanning multiple categories: directly answerable from the structure, requiring inference, and unanswerable
\item Collect Query Agent responses
\item Evaluate responses for accuracy, groundedness, and appropriate uncertainty expression
\end{enumerate}

\subsubsection{Metrics}

We assess query integrity using:
\begin{itemize}
\item \textbf{Answer Accuracy}: Correctness of responses to answerable questions
\item \textbf{Hallucination Rate}: Frequency of fabricated information in responses
\item \textbf{Citation Accuracy}: Correctness of attributions and references in responses
\item \textbf{Uncertainty Calibration}: Appropriateness of confidence expressions and acknowledgment of knowledge gaps
\item \textbf{Source Conflation}: Frequency of incorrectly combining information from distinct sources
\end{itemize}

\subsection{Experimental Design Considerations}

Our evaluation methodology addresses several important experimental design considerations:

\textbf{Domain Diversity}: We evaluate across multiple communication domains to assess generalizability. Initial experiments focus on academic papers due to well-defined structure and verifiable ground truth, but we extend to business communications and technical documentation.

\textbf{Complexity Variation}: We include communication samples ranging from simple (single-topic emails) to complex (multi-section research papers) to understand how trustworthiness scales with information complexity.

\textbf{LLM Model Comparison}: We evaluate multiple LLM backends (GPT-4, Claude, Llama) to understand whether trustworthiness challenges are model-specific or systematic across architectures.

\textbf{Human Expert Validation}: All automated metrics are complemented by human expert evaluation, particularly for semantic fidelity and hallucination detection where automated assessment is challenging.

\section{Implementation: Academic Paper Development}
\label{sec:implementation}

To demonstrate the LAAC framework and enable systematic trustworthiness evaluation, we implemented a complete system for academic paper development and review. This domain provides an ideal testbed: academic papers have well-defined structure, verifiable factual content, and high stakes for information fidelity.

\subsection{System Architecture}

\begin{figure}[htbp]
\centering
\includegraphics[width=0.9\columnwidth]{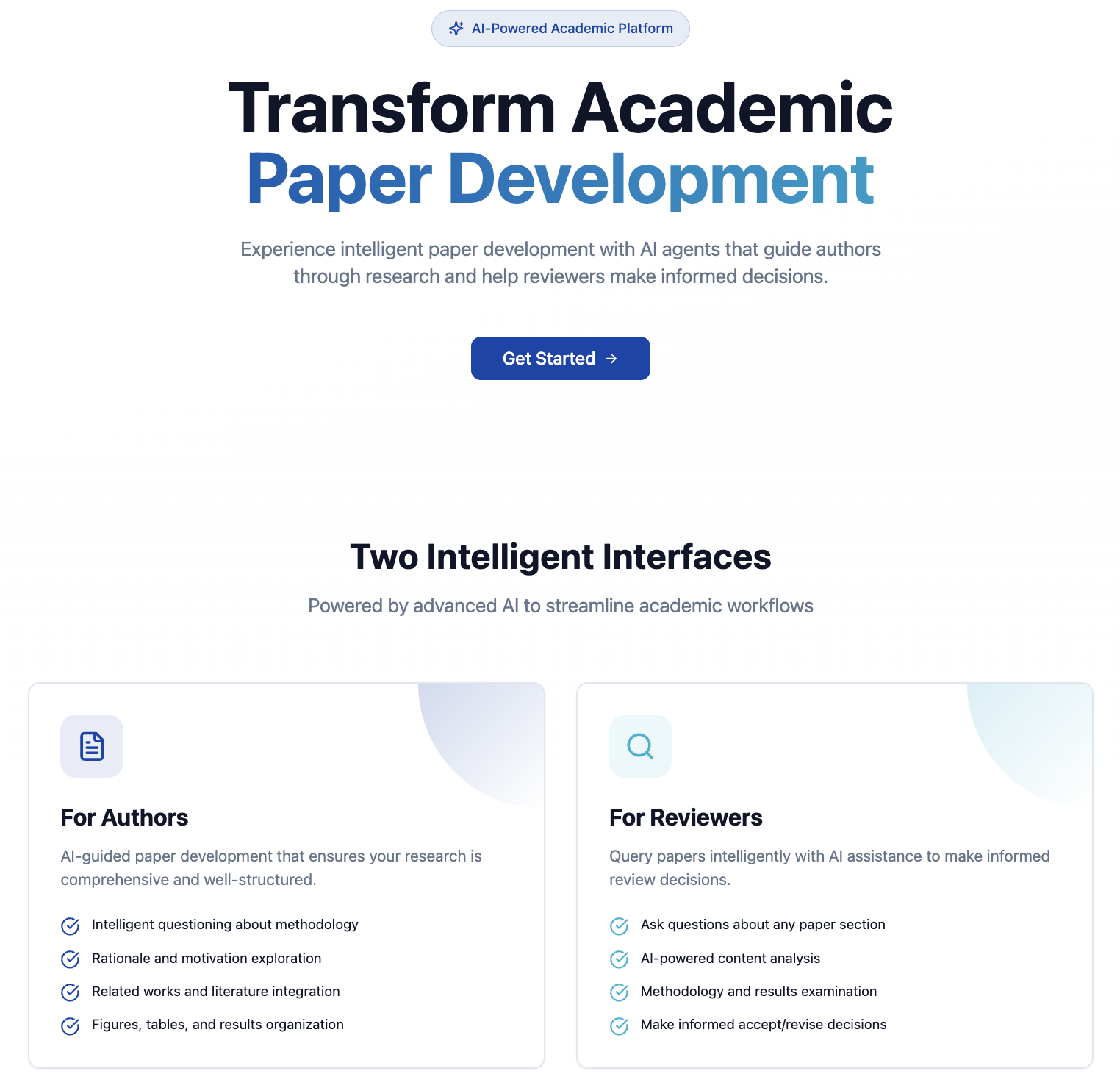}
\caption{LAAC landing page showing dual interfaces for authors and reviewers. The system emphasizes intelligent paper development and AI-powered content analysis to facilitate authentic academic communication.}
\label{fig:landing}
\end{figure}

Figure \ref{fig:landing} shows the landing page of our implementation, which supports two distinct user roles: authors who develop papers through AI-guided interviews, and reviewers who query papers for informed decision-making.

\begin{figure}[htbp]
\centering
\includegraphics[width=0.9\columnwidth]{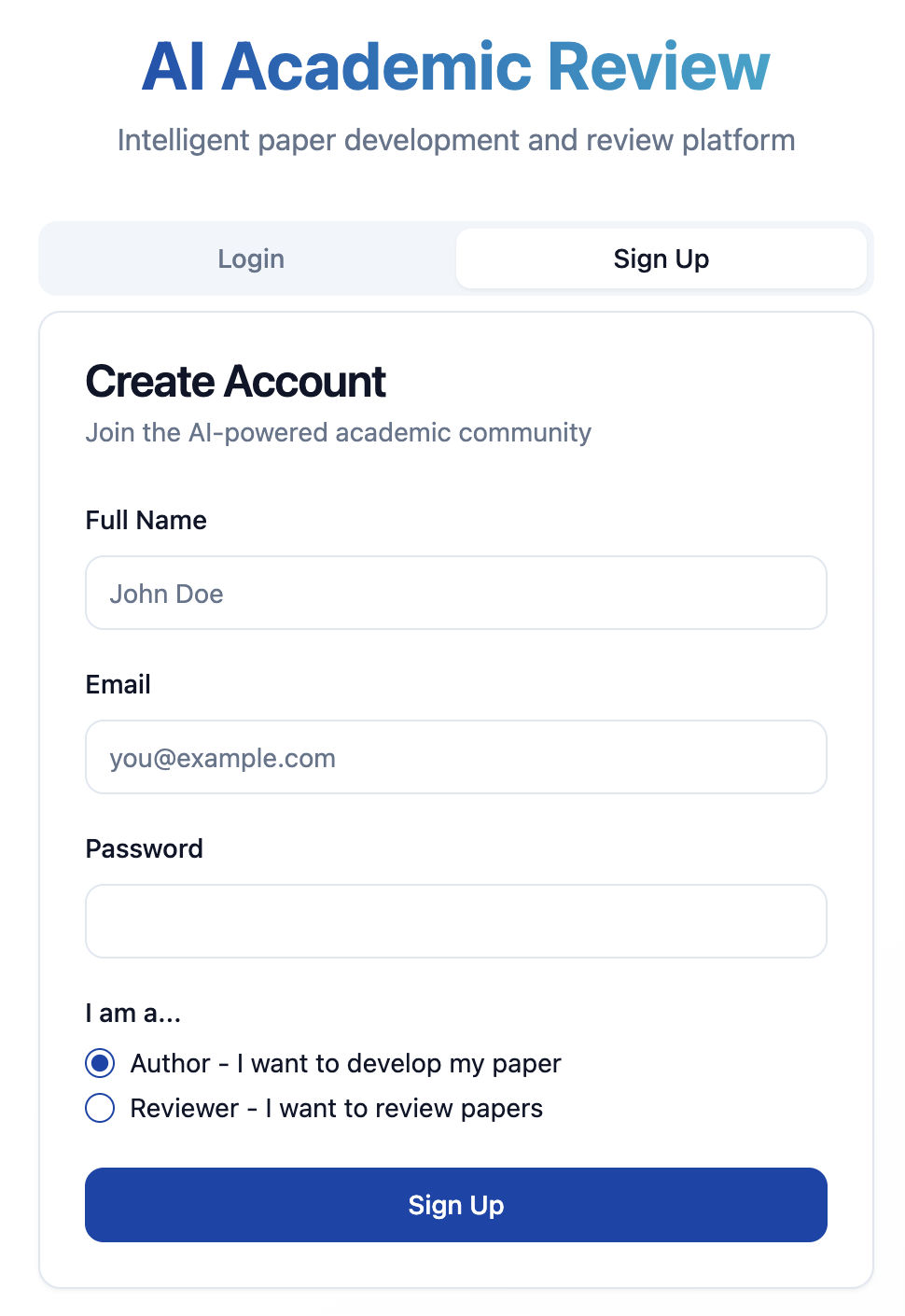}
\caption{Login interface supporting both author and reviewer account types. Role-based access control ensures appropriate permissions for paper development versus review activities.}
\label{fig:login}
\end{figure}

Figure \ref{fig:login} presents the authentication interface, providing access control for both author and reviewer accounts. The system maintains separate workflows optimized for each role while sharing the underlying knowledge representation.

\subsection{Author Interface}

\begin{figure}[htbp]
\centering
\includegraphics[width=0.9\columnwidth]{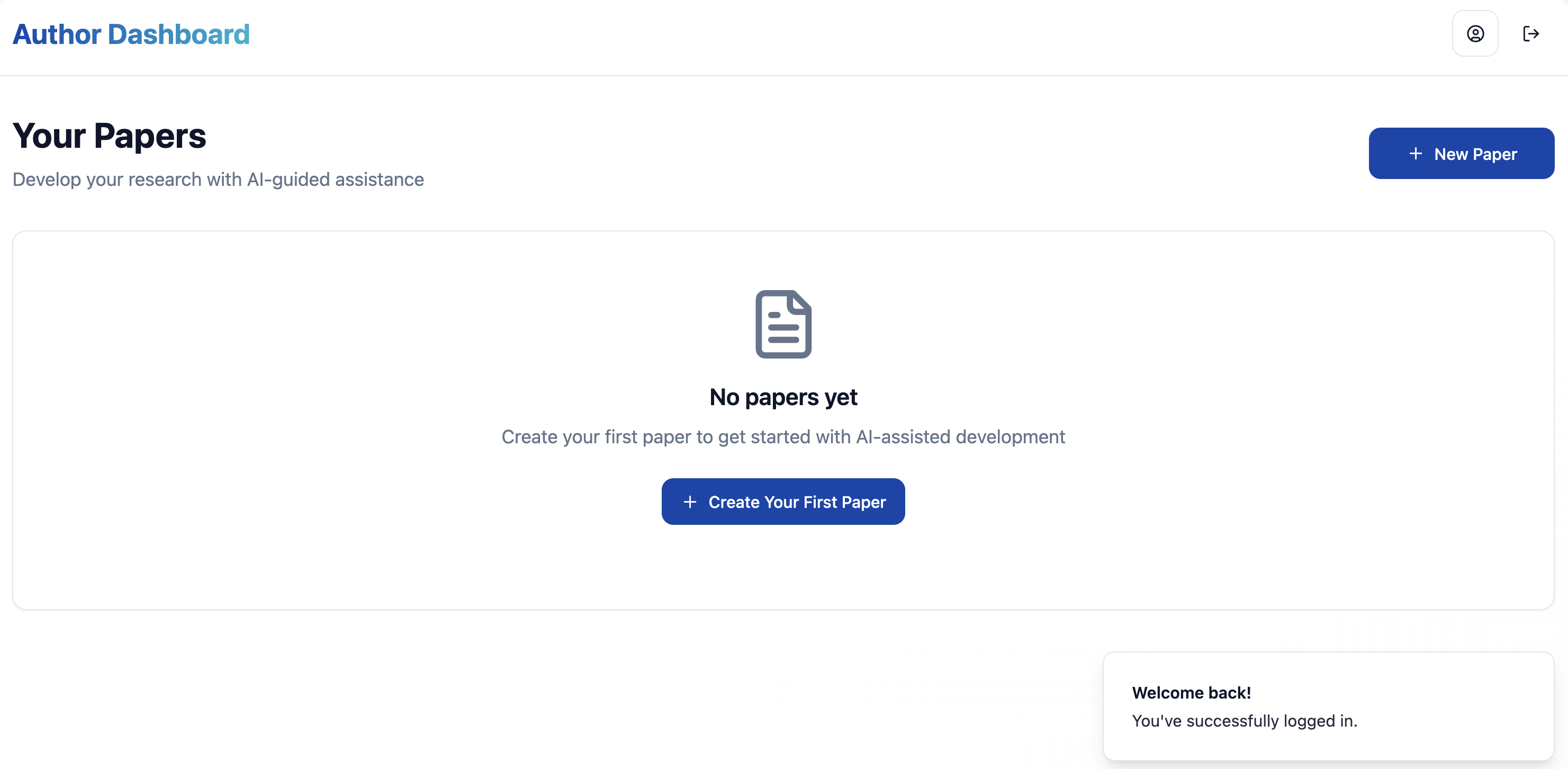}
\caption{Author dashboard for managing papers under development. Authors can create new papers and resume work on existing drafts, with all papers backed by structured knowledge extracted from AI-guided interviews.}
\label{fig:author_dash}
\end{figure}

The author workflow guides researchers through structured paper development via Interview Agent dialogue. Figure \ref{fig:author_dash} shows the author dashboard where researchers manage their papers.

The Interview Agent asks domain-specific questions spanning:
\begin{itemize}
\item Research motivation and problem statement
\item Related work and theoretical positioning
\item Methodology and experimental design
\item Results and key findings
\item Discussion and implications
\item Limitations and future work
\end{itemize}

Authors provide information conversationally, and the system extracts structured knowledge while maintaining an interview transcript. This approach ensures comprehensive coverage while remaining natural and efficient compared to traditional paper writing.

Once the interview is complete, the Extraction Agent generates a hierarchical knowledge structure organized by paper sections. Authors review this structure and can request modifications or provide additional information through follow-up dialogue.

\subsection{Reviewer Interface}

\begin{figure}[htbp]
\centering
\includegraphics[width=0.9\columnwidth]{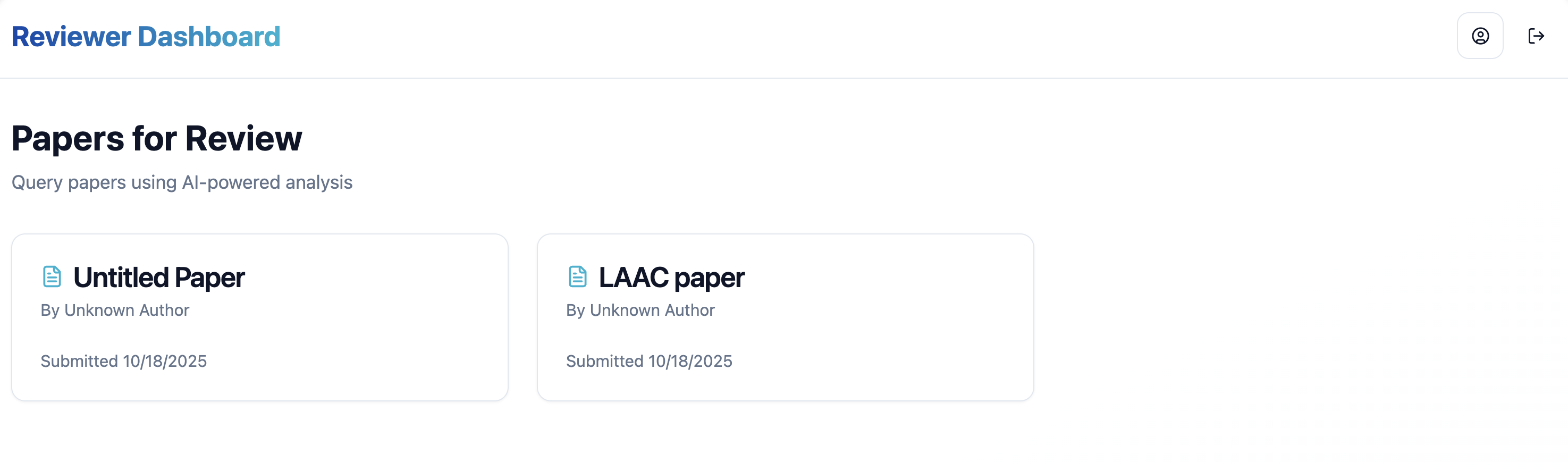}
\caption{Reviewer dashboard showing papers available for review. Reviewers can select papers and engage with the Query Agent to examine methodology, results, and other paper aspects without reading verbose manuscript text.}
\label{fig:reviewer_dash}
\end{figure}

Reviewers interact with submitted papers through the Query Agent. Figure \ref{fig:reviewer_dash} shows the reviewer dashboard displaying available papers.

Rather than reading a traditional manuscript, reviewers ask questions about the research:
\begin{itemize}
\item What methodology did the authors use?
\item What were the primary results?
\item How does this work compare to existing approaches?
\item What are the limitations of the study?
\item Are the claims adequately supported?
\end{itemize}

The Query Agent responds based on the structured knowledge, citing specific sections and providing direct answers. This enables efficient, focused review while ensuring reviewers have access to all relevant information.

\subsection{Technical Implementation}

The system is implemented as a web application with the following components:

\textbf{Frontend}: React-based interface with separate workflows for authors and reviewers. Real-time chat interface for Interview Agent and Query Agent interactions.

\textbf{Backend}: Node.js API server managing user authentication, paper storage, and LLM orchestration. PostgreSQL database for user accounts and paper metadata.

\textbf{LLM Integration}: API integration with Claude (Anthropic) for all three agent roles. Specialized prompts for each agent optimized for their specific functions. Structured knowledge stored as JSON with hierarchical organization.

\textbf{Evaluation Infrastructure}: Logging system capturing all interview transcripts, extracted knowledge structures, and query interactions. Automated metrics calculation and human evaluation interface for trustworthiness assessment.

\section{Preliminary Findings and Discussion}
\label{sec:findings}

We conducted initial experiments evaluating the three trustworthiness dimensions using our academic paper implementation. While comprehensive results require extensive testing across domains, these preliminary findings reveal important challenges and patterns.

\subsection{Information Capture Fidelity Observations}

Testing with five published computer science papers, we observed:

\textbf{Strengths}: The Interview Agent successfully extracted most major paper components (research questions, methodology, results). When authors provided explicit information during interviews, the Extraction Agent generally represented it accurately in the knowledge structure.

\textbf{Weaknesses}: We identified concerning gaps in specific areas:
\begin{itemize}
\item \textbf{Quantitative Results}: Numerical findings were sometimes approximated or omitted. An author stating "we achieved 94.3\% accuracy" might be extracted as "high accuracy was achieved" with the specific number lost.
\item \textbf{Technical Details}: Fine-grained methodological specifics were often summarized at a higher level of abstraction, potentially losing critical reproducibility information.
\item \textbf{Citation Context}: When authors mentioned related work during interviews, the extracted knowledge sometimes misrepresented the relationship (e.g., conflating work being built upon versus work being contrasted against).
\end{itemize}

\textbf{Variable Extraction Accuracy}: Different paper sections exhibited different fidelity levels. Abstracts and high-level contributions were captured with approximately 85-90\% semantic accuracy (human-evaluated), while detailed methodology and statistical results showed 60-70\% fidelity.

\subsection{Reproducibility Challenges}

We processed the same interview transcript through the Extraction Agent ten times with varied temperature parameters:

\textbf{Structural Variability}: Knowledge structures showed significant organizational differences. While major sections (Introduction, Methodology, Results) were consistent, subsection organization varied considerably. One extraction might group related concepts together, while another might distribute them across multiple sections.

\textbf{Detail Inconsistency}: Specific claims appeared, disappeared, or were reworded across extractions. For example, a limitation mentioned in the interview appeared in the extracted knowledge in 7 out of 10 runs—an unacceptable inconsistency level for a communication intermediary.

\textbf{Implicit Information Generation}: We observed instances where the Extraction Agent added implicit connections or implications not explicitly stated in the interview. While sometimes these inferences were reasonable, they represent a departure from strict fidelity that could mislead recipients.

\textbf{Temperature Sensitivity}: Reproducibility strongly correlated with LLM temperature parameter. Lower temperatures (0.1-0.3) produced more consistent structures but sometimes missed nuanced information. Higher temperatures (0.7-1.0) captured more detail but showed unacceptable variability across runs.

\subsection{Query Response Integrity Findings}

Testing with 50 questions across answerable, inference-requiring, and unanswerable categories revealed:

\textbf{Answerable Questions}: For questions directly addressing information in the knowledge structure, the Query Agent provided accurate responses 82\% of the time. The remaining 18\% included minor inaccuracies or omissions.

\textbf{Hallucination Rate}: Critically, when asked unanswerable questions (information not in the knowledge structure), the Query Agent fabricated plausible responses 31\% of the time rather than acknowledging the knowledge gap. This rate is unacceptably high for a trusted intermediary.

\textbf{Citation Fabrication}: When the knowledge structure contained citations, the Query Agent sometimes invented specific page numbers, publication years, or author names that were not provided in the original information—a particularly concerning form of hallucination for academic communication.

\textbf{Source Conflation}: In papers discussing multiple related works, the Query Agent occasionally attributed findings from one paper to another, or combined results from multiple studies when answering comparative questions.

\textbf{Overconfidence}: The Query Agent rarely expressed uncertainty even when making inferences beyond the explicit knowledge structure. This lack of calibrated confidence is problematic for recipients trying to assess information reliability.

\subsection{Cross-Cutting Observations}

Several patterns emerged across all three trustworthiness dimensions:

\textbf{Information Loss at Agent Boundaries}: Each agent transition (Interview to Extraction, Extraction to Query) introduced information loss or distortion. The cumulative effect across the pipeline means recipient-facing information may diverge significantly from sender intent.

\textbf{Abstraction Tendency}: The system consistently pushed toward higher levels of abstraction, losing specific details. This may reflect LLM training on summarization tasks, but it conflicts with LAAC's goal of maintaining full information fidelity.

\textbf{Domain Formality Bias}: The system showed bias toward formal academic language patterns even when authors provided information conversationally. This could obscure author intent behind generic academic phrasing.

\textbf{Lack of Uncertainty Tracking}: The knowledge structure format did not adequately distinguish between explicitly stated facts, reasonable inferences, and uncertain claims. All information was presented with equal confidence in query responses.

\subsection{Implications for LAAC Deployment}

These preliminary findings suggest that current LLM technology, while powerful, exhibits systematic trustworthiness gaps that must be addressed before LAAC systems can be deployed as reliable communication intermediaries:

\begin{enumerate}
\item \textbf{Fidelity Verification Mechanisms}: Systems need explicit verification steps where senders confirm the accuracy of extracted knowledge before recipients access it.
\item \textbf{Provenance Tracking}: Knowledge structures should maintain direct links to specific interview statements, enabling recipients to verify information sources.
\item \textbf{Uncertainty Quantification}: Systems must explicitly represent and communicate confidence levels for different pieces of information.
\item \textbf{Hallucination Detection}: Query Agents require robust mechanisms for detecting when questions cannot be answered from available knowledge and refusing to speculate.
\item \textbf{Human-in-the-Loop Validation}: For high-stakes communication domains, automated extraction should be complemented by human review and approval.
\end{enumerate}

\section{Limitations and Future Work}

This position paper presents preliminary work with several important limitations that motivate future research directions.

\subsection{Current Limitations}

\textbf{Limited Experimental Scope}: Our evaluation focused primarily on academic papers with a small sample size. Comprehensive assessment requires larger datasets spanning diverse communication domains, complexity levels, and LLM models.

\textbf{Single Implementation}: We evaluated one specific implementation of the LAAC architecture. Different design choices (agent prompts, knowledge schemas, interaction patterns) may yield different trustworthiness profiles.

\textbf{Preliminary Metrics}: Our evaluation metrics, while systematically defined, have not been validated against user trust perceptions or real-world deployment outcomes. The relationship between measured fidelity and perceived trustworthiness requires investigation.

\textbf{Controlled Experiments}: We evaluated using structured protocols with researcher-controlled inputs. Real-world usage with diverse users may reveal additional failure modes not captured in controlled settings.

\subsection{Future Research Directions}

\textbf{Advanced Trustworthiness Mechanisms}: We plan to investigate technical approaches for improving LAAC trustworthiness including retrieval-augmented extraction that grounds knowledge structures in interview transcripts, uncertainty-aware query responses that explicitly acknowledge knowledge boundaries, multi-model consensus where multiple LLMs independently extract knowledge and their outputs are reconciled, and formal verification methods that provide guarantees about information fidelity for specific properties.

\textbf{User Studies}: Comprehensive evaluation requires understanding how real users perceive LAAC system trustworthiness. We plan controlled studies comparing LAAC-mediated communication to traditional methods, investigating user trust calibration, and identifying domain-specific trustworthiness requirements from practitioners.

\textbf{Domain Expansion}: We will extend LAAC implementation and evaluation to additional communication domains including business proposals, technical documentation, grant applications, policy briefs, and cross-cultural communication scenarios.

\textbf{Standardized Benchmarks}: The broader research community would benefit from standardized benchmarks for evaluating communication intermediary systems, enabling rigorous comparison across approaches and tracking progress over time.

\textbf{Regulatory and Ethical Considerations}: As LAAC-style systems become more prevalent, important questions arise about liability when AI intermediaries misrepresent information, disclosure requirements for AI-mediated communication, and standards for acceptable trustworthiness levels in different domains.

\section{Conclusion}
\label{sec:conclusion}

The proliferation of LLMs has created a paradoxical situation where these powerful tools are being used to inflate and then compress information, wasting computational resources while eliminating authentic human communication. The LAAC framework offers an alternative paradigm: positioning LLMs as trusted intermediaries that faithfully capture sender intent and facilitate recipient understanding through structured knowledge and interactive querying.

However, our systematic evaluation reveals that deploying LLMs in this communication intermediary role faces critical trustworthiness challenges. Current systems exhibit measurable gaps in information capture fidelity, reproducibility, and query response integrity. Interview Agents achieve variable extraction accuracy, Extraction Agents produce inconsistent knowledge structures from identical inputs, and Query Agents show concerning tendencies toward hallucination and citation fabrication.

These findings should not be interpreted as fundamental limitations of the LAAC approach, but rather as empirical validation that trustworthiness cannot be assumed—it must be systematically evaluated and engineered. The path forward requires technical innovations in fidelity verification, uncertainty quantification, and hallucination prevention, combined with appropriate human-in-the-loop validation for high-stakes communication domains.

As LLMs continue to evolve, their potential role as communication intermediaries represents a more authentic and efficient application than content generation. However, realizing this potential requires the AI research community to prioritize trustworthiness alongside capability, developing systems that users can rely on to faithfully represent their intent. This paper provides a framework for understanding and measuring that trustworthiness, establishes preliminary baselines, and charts a path toward reliable LLM-mediated communication.

\bibliographystyle{plain}
\bibliography{references}

\end{document}